\documentclass[letterpaper, 10 pt, conference]{ieeeconf}  

\usepackage{mathtools}
\usepackage{multirow}
\usepackage{graphics}
\usepackage{subfig}
\usepackage{float}
\usepackage{stfloats}
\usepackage{times}
\usepackage{epsfig}
\usepackage{graphicx}
\usepackage{amsmath}
\usepackage{amssymb}
\usepackage{todonotes}
\usepackage{hyperref}
\usepackage{url}
\graphicspath{ {figures/} }
\newcommand{\OF}{OF}
\newcommand{\shorteq}{%
  \settowidth{\@tempdima}{-}
  \resizebox{\@tempdima}{\height}{=}%
}

\IEEEoverridecommandlockouts                              

\title{
\bf
VOLDOR$^+$SLAM: \\
\resizebox{2\columnwidth}{!}
{For the times when feature-based or direct methods are not good enough}
}

\author{Zhixiang Min$^{1}$ and Enrique Dunn$^{1}$
\thanks{$^{1}$Stevens Institute of Technology {\tt\small \{zmin1,edunn\}@stevens.edu}}%
}

\begin{document}

\maketitle
\thispagestyle{empty}
\pagestyle{empty}

\begin{abstract}
We present a dense-indirect SLAM system using external dense optical flows as input.
We extend the recent probabilistic visual odometry model VOLDOR \cite{min2020voldor}, by incorporating the use of geometric priors to 1) robustly bootstrap estimation from monocular capture, while  2) seamlessly supporting stereo and/or RGB-D input imagery.
Our customized back-end tightly couples our intermediate geometric estimates with an adaptive priority scheme managing the connectivity of an incremental pose graph. 
We leverage recent advances in dense optical flow methods to achieve accurate and robust camera pose estimates, while constructing fine-grain globally-consistent dense environmental maps. 
  Our open source implementation {\color{purple} [https://github.com/htkseason/VOLDOR]} operates online at around 15 FPS on a single GTX1080Ti GPU.
\end{abstract}


\section{Introduction}
Simultaneous localization and mapping (SLAM) addresses the incremental construction and instantaneous establishment of a global geometric reference using  measurements from a dynamic observer as input.  Applications benefiting from such capabilities include robotics, augmented/virtual reality, and autonomous driving. 
For visual SLAM, the choice of geometric representation and estimation framework  
must balance requirements for  online operation  (e.g. latency and computational burden) and  achievement of task-relevant performance metrics (e.g. robustness, accuracy, and completeness). Recent dense-indirect methods  challenge the standard dichotomy  among feature-based and direct SLAM methods. 

\begin{figure}[t!]
\centering
\includegraphics[width=\columnwidth]{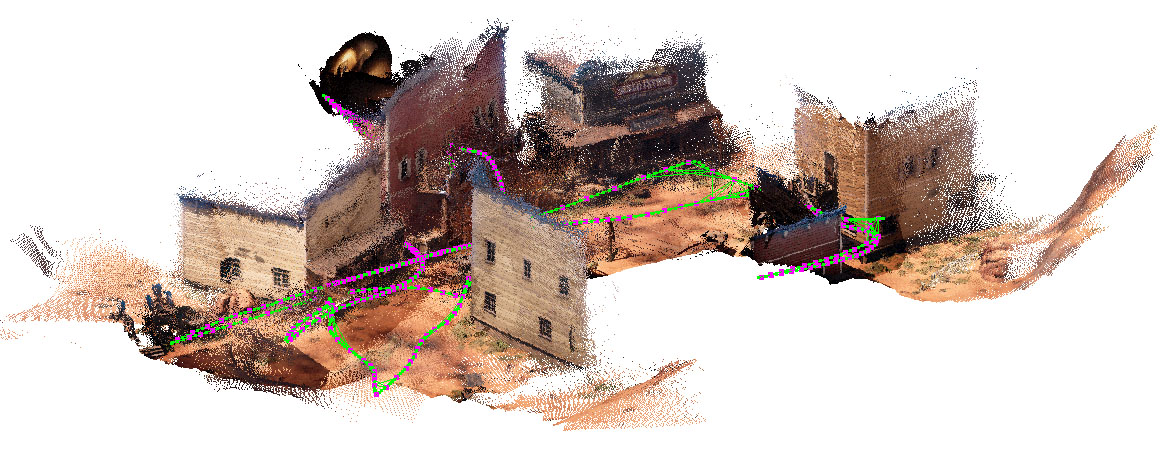}
\text{ \small (a) Result on TartanAir 'westerndesert' sequence with stereo input.}
\includegraphics[width=0.5\columnwidth]{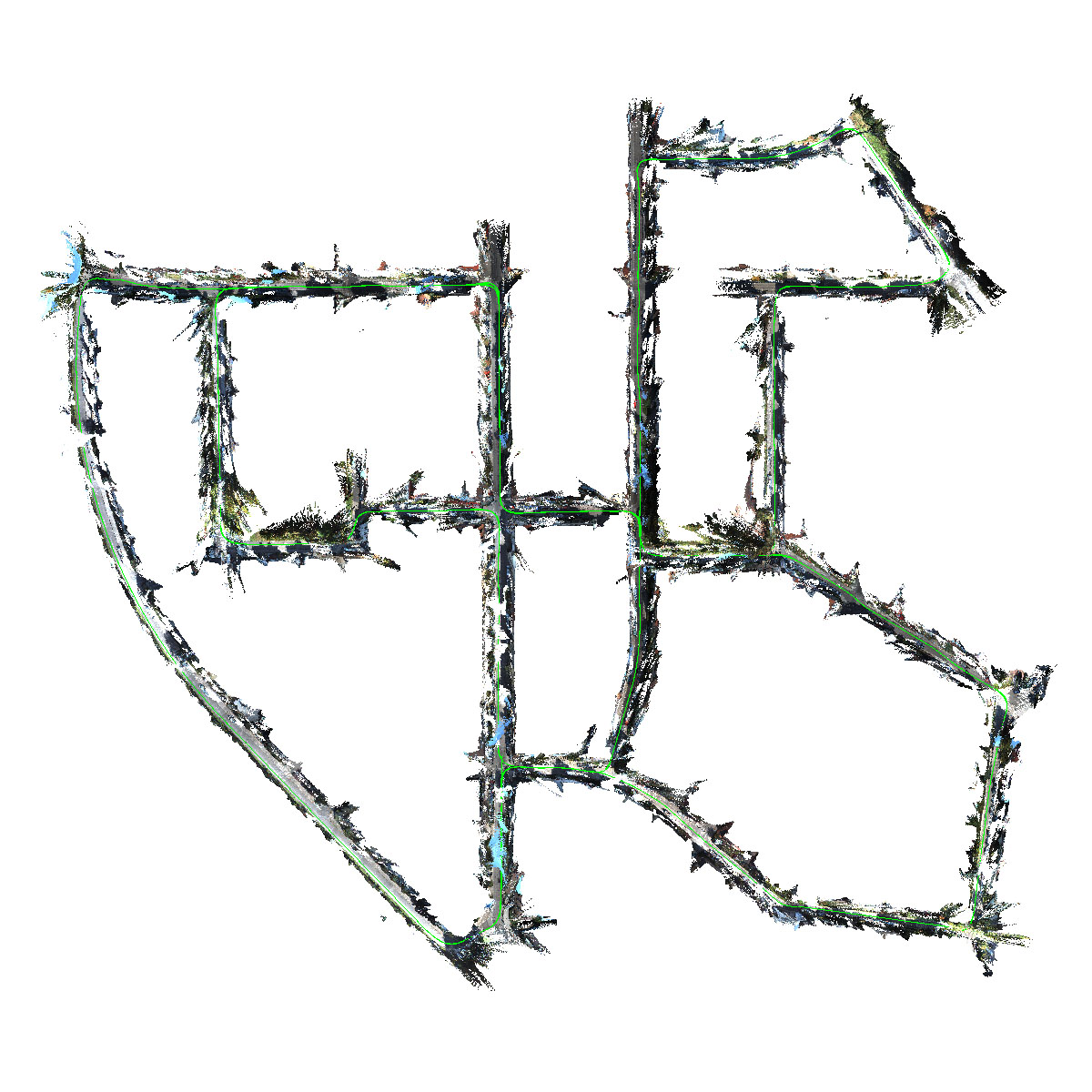}
\includegraphics[width=0.48\columnwidth]{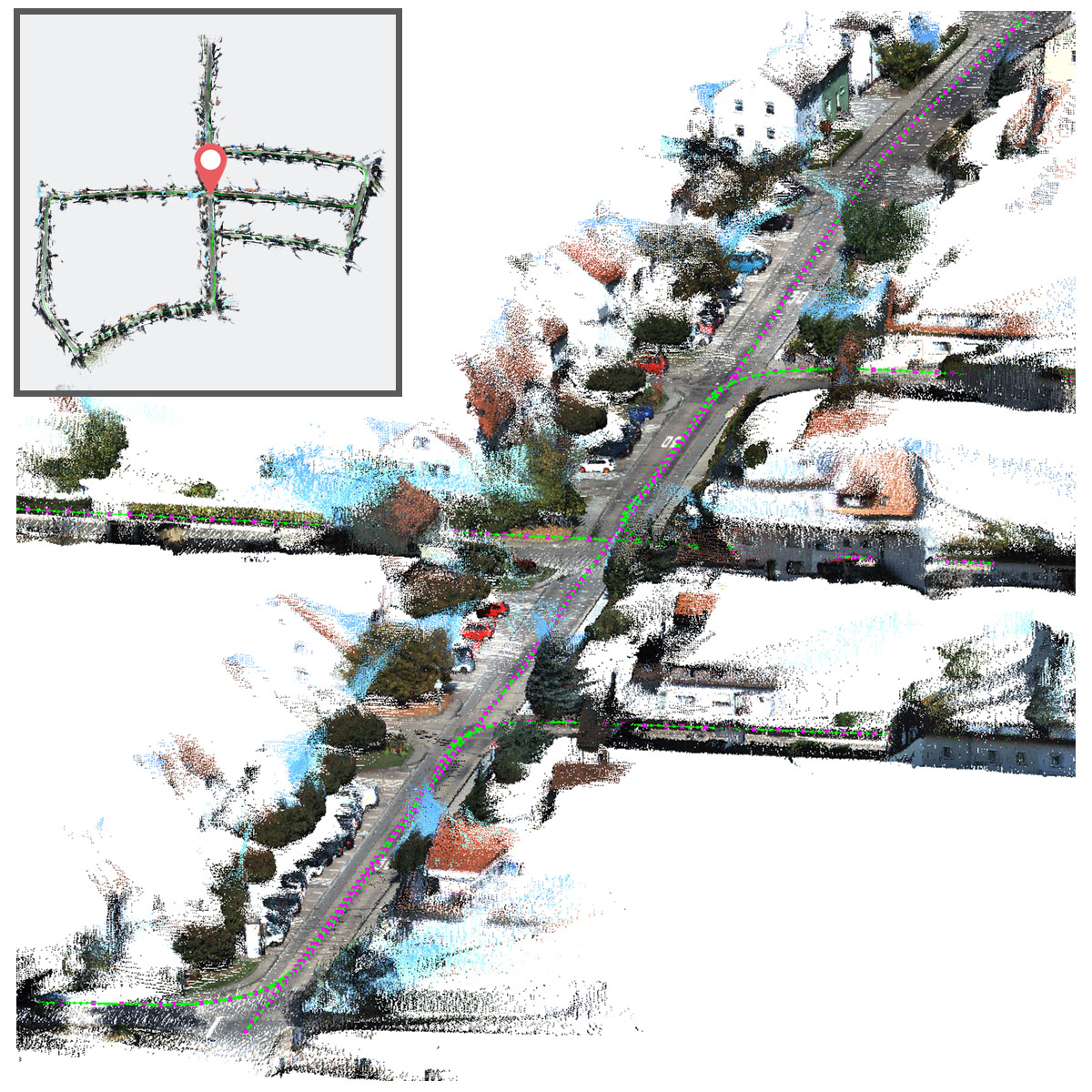}
\text{\small (b) Result on KITTI 00, 05 sequence with stereo input.}
\caption{Reconstructed scene models.  3D point clouds are aggregated from keyframe depth maps. }
\label{fig:title_figure}
\end{figure}

Feature-based methods determine multi-view relationships among input video frames through the geometric analysis of sparse key-point correspondences.  Popular feature-based SLAM systems \cite{pire2015stereo, mur2017orb} have a visual odometry (VO) front-end solving feature correspondence/triangulation and PnP instances to initialize camera pose estimates, which are subsequently refined (e.g.  bundle-adjusted) by back-end and mapping modules. Given that sparse feature analysis can operate on real-time, the trade-offs between online operation and  accuracy/robustness achieved by these methods is mainly determined by the scope of the analysis performed by the back-end. However, the accuracy of  front-end  modules may be affected by lack of texture content, repetitive structures, and/or degenerate geometric configurations; compromising the efficacy and efficiency of back-end modules.

Direct methods 
jointly estimate photometrically consistent scene structure and camera poses from image content, obviating key-point correspondence. Dense direct methods \cite{6126513} are typically sensitive to illumination changes, and need a sufficient number of images to triangulate structure at poorly textured regions. Sparse or semi-dense direct methods \cite{engel2014lsd, engel2017direct} are less sensitive to illumination changes, but may still require accurate photometric calibration for full performance.

Recent dense-indirect (DI) formulations for VO \cite{min2020voldor} are a promising alternative to both their sparse and direct counterparts. The DI approach conditions local 3D geometry and camera motion estimates on their consistency w.r.t. (observed) dense optical flow (OF) estimates. Such frameworks actively leverage ongoing improvements in accuracy and robustness of learning-based \OF{} estimators, while yielding high-quality dense geometry estimates as a byproduct.

To extend the recent DI 
inference model VOLDOR \cite{min2020voldor} into a SLAM pipeline,  we develop integrative modules customized to the intermediate geometric estimates produced by its inference process. 
We address the efficient estimation, management and refinement of global-scope 3D data associations within the context of DI geometric estimates. Our contributions are:
1) extending VOLDOR's inference framework to both generate and input explicit geometric priors for improved estimation accuracy across monocular, stereo and RGB-D image sources,
2) a robust point-to-plane inverse-depth alignment formulation for source-agnostic frame registration, and 3) a priority linking framework for incremental real-time pose graph management. 

\section{Related Work}

\noindent \textbf{Indirect SLAM}. Classic approaches like MonoSLAM \cite{davison2007monoslam} rely on EKF-based frameworks for fusing multiple observations to recover scene geometry. PTAM \cite{klein2007parallel}  simultaneously runs a front-end for VO and a back-end refining the estimation through bundle adjustment. ORB-SLAM \cite{mur2015orb, mur2017orb}, introduced a versatile SLAM system with a more powerful back-end with global re-localization and loop closing, allowing large environment applications. More recently, ORB-SLAM3 \cite{campos2020orb} further built a multi-map system to survive from long periods of poor visual information.  

\noindent \textbf{Direct SLAM}. DTAM \cite{6126513} first introduced a GPU-based real-time dense modelling and tracking approach for small workspaces. It estimates a dense geometry model through a photometric term combined with a regularizer while estimating  camera motion by finding a warping of the model that minimizes the photometric error w.r.t. the video frames. LSD-SLAM \cite{engel2014lsd} switched to keyframe based semi-dense model that allows large scale CPU real-time application. DSO \cite{engel2017direct} builds sparse models and combines a probabilistic model that jointly optimizes for all parameters as well as further integrates a full photometric calibration.

\noindent \textbf{Direct RGB-D SLAM}.
RGB-D SLAM \cite{kerl2013dense} forgoes  geometry estimation by using RGB-D input, while  estimates camera poses by minimizing both photometric and geometric error, and  managing keyframes within a pose graph. 
ElasticFusion  \cite{whelan2015elasticfusion} uses dense frame-to-model camera tracking, windowed surfel-based fusion, and frequent model refinement through non-rigid surface deformation.  BundleFusion  \cite{dai2017bundlefusion} tracks camera poses using sparse-to-dense optimization and a local-to-global strategy for global mapping consistency.
Recently, BAD-SLAM \cite{schops2019bad} proposed a fast direct bundle-adjustment for poses, surfels and intrinsics.

\noindent \textbf{Dense-Indirect VO.} Valgaerts et al. \cite{valgaerts2012dense} addressed robust recovery of fundamental matrix from a single dense \OF{} field. Ranftl et al. \cite{ranftl2016dense} further addressed the motion segmentation and the recovery of depth maps. While 
these works highlighted the potential of DI modeling, they focused on the two-view geometry problem. VOLDOR \cite{min2020voldor} introduced the DI paradigm to the multi-view domain by modelling the VO problem within a probabilistic framework w.r.t. a log-logistic residual model on \OF{} batches and developed a generalized-EM framework for joint real-time inference of camera motion, 3D structure and track reliability. While highly accurate,  geometric analysis scope was strictly local.

\noindent \textbf{Deep learning optical flow.} Current deep learning works on \OF{} estimation outperform traditional methods in terms of accuracy and robustness under challenging conditions such as texture-less regions, motion blur and large occlusions. FlowNet/FlowNet2 \cite{dosovitskiy2015flownet, ilg2017flownet} introduced an encoder-decoder CNN for \OF{} estimation. PWC-Net \cite{sun2018pwc} integrates spatial pyramids, warping and cost volumes into deep \OF{} estimation, improving performance and generalization. Most recently, MaskFlowNet \cite{zhao2020maskflownet} proposed an asymmetric occlusion aware feature matching module and achieved current state-of-the-art performance.


\begin{figure*}[t]
\centering
\includegraphics[width=2.05\columnwidth]{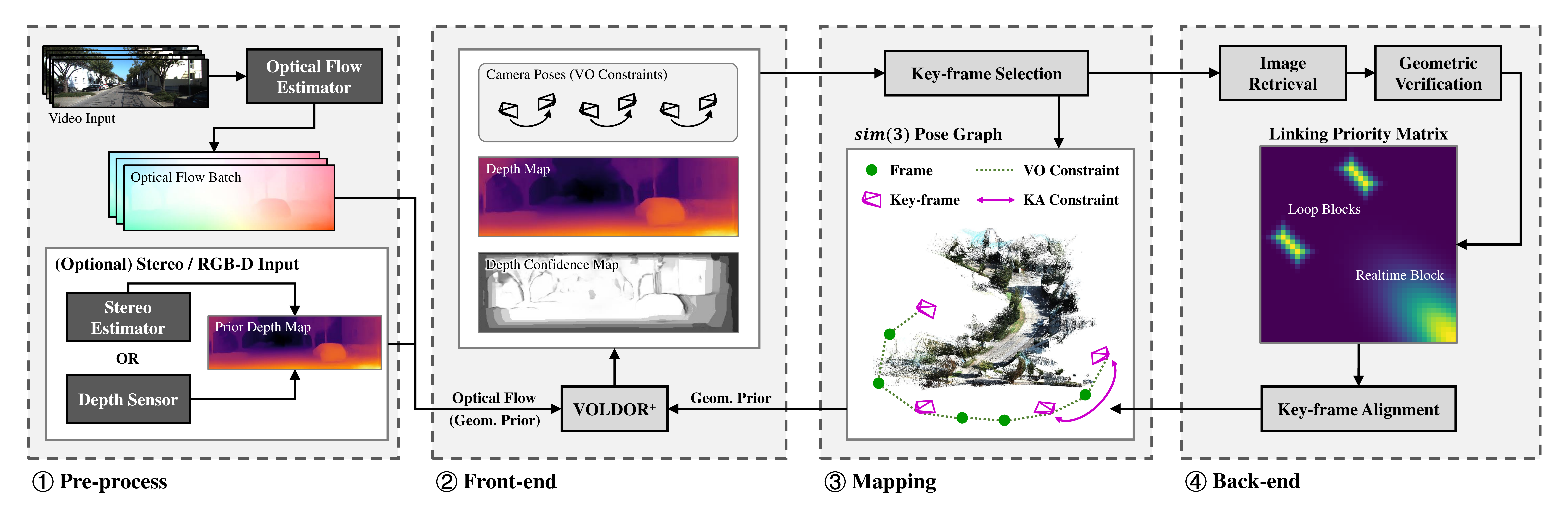}
\caption{VOLDOR$^+$SLAM architecture.
1)  Input optical flow is externally  computed  from  video, along with optional geometric priors. 
2) Dense-indirect VO front-end estimates scene structure and local camera poses over a sliding window.
3) A pose graph enforces global consistency among all pairwise pose estimates. 4) The set of edges to include in our pose graph is prioritized based on keyframe geometric analysis aimed at both identifying loop closures and reinforce local connectivity. 
}

 
\label{fig:workflow}
\end{figure*}

\section{Method}

\noindent {\bf System Architecture}. We build upon the recently proposed dense indirect VO method of Min et al. 
\cite{min2020voldor}, which addressed the joint probabilistic estimation of camera motion, 3D structure and track reliability from a set of input dense \OF{} estimates.  As standard practice in SLAM literature \cite{mur2017orb, engel2017direct},  our proposal VOLDOR$^+$SLAM has a VO front-end and a mapping back-end. The front-end operates over small sequential batches (i.e. temporal sliding window) of dense \OF{} estimates. We adaptively determine keyframe selection and stride among subsequent batches based on visibility-coverage metrics designed for our dense SLAM system. To enforce consistency within a larger geometric scope, while enabling online operation, the back-end  adaptively prioritizes the analysis and establishment pose constraints between keyframe pairs balancing the search for loop closure connections and the reinforcement of local keyframe connectivity. VOLDOR$^+$SLAM also implements a loop closure scheme based on an image retrieval and geometric verification module \cite{GalvezTRO12}. Finally, all pairwise camera pose constraints are managed within a $sim(3)$-based pose graph \cite{strasdat2010scale}. Module dependencies and data flow of our system are illustrated in Fig.\ref{fig:workflow}. The resulting dense SLAM implementation enables online operation at $\sim\!\!15$ FPS on a single GTX1080Ti GPU.

\subsection{Front-End} \label{sec:front_end}
\noindent{\bf VOLDOR$^+$: An extended VO model}.
Our front-end extends and improves upon the VOLDOR framework \cite{min2020voldor},
which formulates monocular VO from observed \OF{} input batches as a generalized EM problem where the three hidden variables are camera pose, depth, and track reliability.    Originally, \cite{min2020voldor} proposed the probabilistic graphical model 
\begin{equation} \label{eq:voldor_prob_model}
    P(\mathbb{X} \;|\; \mathbb{T}, \boldsymbol{\theta}, \mathbb{W})
\end{equation}
where $\mathbb{X} \text{=} \{ \boldsymbol{X_t} \,|\, t\text{=}1,\! \cdots \!, N_t \}$ denotes the observed \OF{} batch, $\mathbb{T} \text{=} \{ \boldsymbol{T_t} \,|\, t\text{=}1,\! \cdots \!, N_t \}$ the set of camera poses, $\boldsymbol{\theta}$ the depth map of the first frame in the batch and $\mathbb{W} \text{=} \{ \boldsymbol{W_t} \,|\, t\text{=}1,\! \cdots \!, N_t \}$ the set of rigidness maps used to down-weight  \OF{} pixels belonging to occlusions, moving objects or outliers.
The likelihood function in \cite{min2020voldor} defines residuals as the end-point-error (EPE) between the observed optical flow vs. the rigid flow computed from the parameters.
This residual is assumed to follow an empirically validated log-logistic (i.e. Fisk) distribution model. The ensuing MAP problem is solved using a generalized-EM framework  alternatively updating each of the parameters. 
Whereas \cite{min2020voldor} bootstraps each of the hidden variables deterministically, VOLDOR$^+$ accommodates explicit priors, extending  Eq.\eqref{eq:voldor_prob_model} to yield
\begin{equation}
    P(\mathbb{X}, \boldsymbol{\hat{\Theta}} \;|\; \mathbb{T}, \boldsymbol{\theta}, \mathbb{W}, \hat{\mathbb{W}} \;;\; \hat{\mathbb{T}}),
\end{equation}
where the set of explicit depth map priors is denoted by
$\boldsymbol{\hat{\Theta}} \text{=} \{\boldsymbol{\hat{\theta}_k} \,|\, k\text{=}1,\! \cdots \!, N_k\}$; their associated ({\em fixed}) relative camera poses  by $\hat{\mathbb{T}} \text{=} \{ \boldsymbol{\hat{T}_k} \,|\, k\text{=}1,\! \cdots \!, N_k \}$ and their pixel-level rigidness maps by $\hat{\mathbb{W}} \text{=} \{ \boldsymbol{\hat{W}_k} \,|\, k\text{=}1,\! \cdots \!, N_k \}$.
Henceforth, unless otherwise stated, our VOLDOR$^+$ extensions strictly follow the framework described in \cite{min2020voldor} and we refer the reader to that publication for further details.

\noindent \textbf{Geometric Priors}. While \cite{min2020voldor} estimates depth maps exclusively from monocular \OF{} fields, VOLDOR$^+$ extends the probabilistic model to optionally account for input depth map priors. 
Limiting our assumptions (for now) on the source of these priors, to knowing  their first and second moments, we model a generic likelihood function of the depth value at pixel $j$  as a Gaussian-Uniform mixture
\begin{equation} \label{eq:depth_prior_likelihood}
        P( \hat{\theta}_k^{\ell} \;|\; \theta^j, \! \hat{W}_k^j ;  \hat{\boldsymbol{T}}_{\!k})
        \!=\!\begin{cases}
            \!\mathcal{N}^{\;0}_{\!\sigma \left(\! 1/\hat{\theta}_{k}^{\ell} \!\right)} 
                \!\!\left( \!
                \frac{1}{
                \hat{\theta}_{k}^{\ell} 
                }
                \!-\! 
                \frac{1}{
                \phi_z\!(\!\theta^j\!, \hat{\boldsymbol{T}}_{\!k}\!)
                }
                \!\right) 
                & \text{\!if \!$\hat{W}_k^j\!=\!1$}
                \\
                \!\mathcal{U}\!\!\left(\!\frac{1}{\hat{\theta}_{k}^{\ell}} \!  \right) 
                & \text{\!if\! $\hat{W}_k^j\!=\!0$}
            \end{cases}
\end{equation}
where $\phi_z(\!\theta^j\!,\hat{\boldsymbol{T}}_{\!k}\!)$ denotes the $Z$-component of the camera-space 3D coordinates of a depth value transferred across frames having relative motion $\boldsymbol{T}_{\!k}$, 
$\ell \!\!=\!\! \pi(\boldsymbol{\theta}_2^j, \hat{\boldsymbol{T}}_k)$ denotes the pixel of the reprojection $\pi$ of a depth estimate across camera frames,
 while $\sigma(\cdot)$ and $\mathcal{U}(\cdot)$ are functions adjusting the variance of a Gaussian distribution and the density of a uniform distribution w.r.t. the observed depth prior, which are set in proportion to inverse depth value. As in \cite{min2020voldor}, we apply the likelihood function to a maximum inlier estimation framework to define the energy function for prior depths
\begin{equation}
E_{\text{geom-prior}} = 
    -\!\sum_k \!q(\hat{W}_k^j) \;
         log \scalebox{1.2}{$
            \frac{P( \hat{\theta}_k^{\ell} \;|\; \theta^j, \hat{W}_k^j=1)}
            {\sum_{\hat{W}_k^j} P( \hat{\theta}_k^{\ell} \;|\; \theta^j,  \hat{W}_k^j)}
            $},
\end{equation}
where $q(\hat{W}_k^j)$ is the probability density of $\hat{W}_k^j$.
Finally, aggregating with the original energy function conditioned on input \OF{} described in  \cite{min2020voldor}, optimal depth values are selected based on the criterion
\begin{equation}
    \theta^j_{\text{opt}} = \underset{\theta^{j*}}{argmin} (E_{\text{optical-flow}} + E_{\text{geom-prior}})
\end{equation}
In practice each VO batch will typically have two prior depth maps: One generated during the analysis of the previous batch and another from the most recent available keyframe. Whenever both instances point to the same depth map, we will use that single depth prior. When stereo capture (or RGB-D imagery) is available, we add the corresponding depth map as a prior. However, in such case (i.e. depth priors come from known external source), VOLDOR$^+$ replaces Eq.\eqref{eq:depth_prior_likelihood} by an empirical residual model as done in \cite{min2020voldor}.

\noindent \textbf{Depth Map Confidence Estimation}. 
 For each depth map $\boldsymbol{\theta}$ generated through the VOLDOR$^+$ framework, we associate a confidence map $\boldsymbol{C}$, which we define as the pixel-wise average of previously estimated rigidness of optical flows and depth priors used for its estimation:
\begin{equation} \label{eq:depth_conficence}
    C^j = \frac{\sum_t{W_t^j} + \sum_k{\hat{W}_k^j}}{N_t+N_k}
\end{equation}
When estimated depth maps are subsequently used as depth priors, the value $C^j$ is used as a prior for $\hat{W}_k^j$ in Eq.\eqref{eq:depth_prior_likelihood}, such that a new weight $C^j \hat{W}_k^j$ will replace $\hat{W}_k^j$ in the depth update step. Similarly, $C^j$ is also used in the back-end keyframe alignment as will be described in Eq.\eqref{Eq_KA}.

\noindent{\bf Uncertainty-aware Pose Estimation}. The probabilistic camera pose inference in \cite{min2020voldor} assumes other parameters fixed and approximates the camera pose posterior distribution by Monte-Carlo sampling of P3P instances
\begin{equation}
    P(\mathbb{T} \mid \mathbb{X} \,;\, \boldsymbol{\theta}, \mathbb{W}) 
    \approx \prod_t \left[
    \frac{1}{S}
    \sum_g^S 
    q(\boldsymbol{T}_t^g)
    \right]
\end{equation}
where $t$ denotes the time stamp, $S$ the total number of samples, $g$  the sample index, and $q(\boldsymbol{T}_t^g)$ is a variational distribution approximating the intractable sample posterior. Further, let $q(\boldsymbol{T}_t^g)$ be a normal distribution $\mathcal{N}(\boldsymbol{\hat{T}}_t^g,\boldsymbol{\Sigma})$, where $\boldsymbol{\hat{T}}_t^g$ is solved using a AP3P solver \cite{ke2017efficient}, while the covariance matrix $\boldsymbol{\Sigma}$ is a fixed hyper-parameter. The camera pose defined by the mode of the approximated posterior, is found using meanshift with a Gaussian kernel in Lie algebra.
To accommodate subsequent back-end modules, VOLDOR$^+$ estimates each camera pose uncertainty $\boldsymbol{\Sigma}_t$ by using the meanshift result as initialization and iteratively fitting a Gaussian distribution to the samples. We discard samples outside 3-sigma for robustness and force all eigenvalues of $\boldsymbol{\Sigma}_t$ to be larger than an epsilon to ensure numerical stability.

\begin{figure}[]
\centering
\includegraphics[width=\columnwidth]{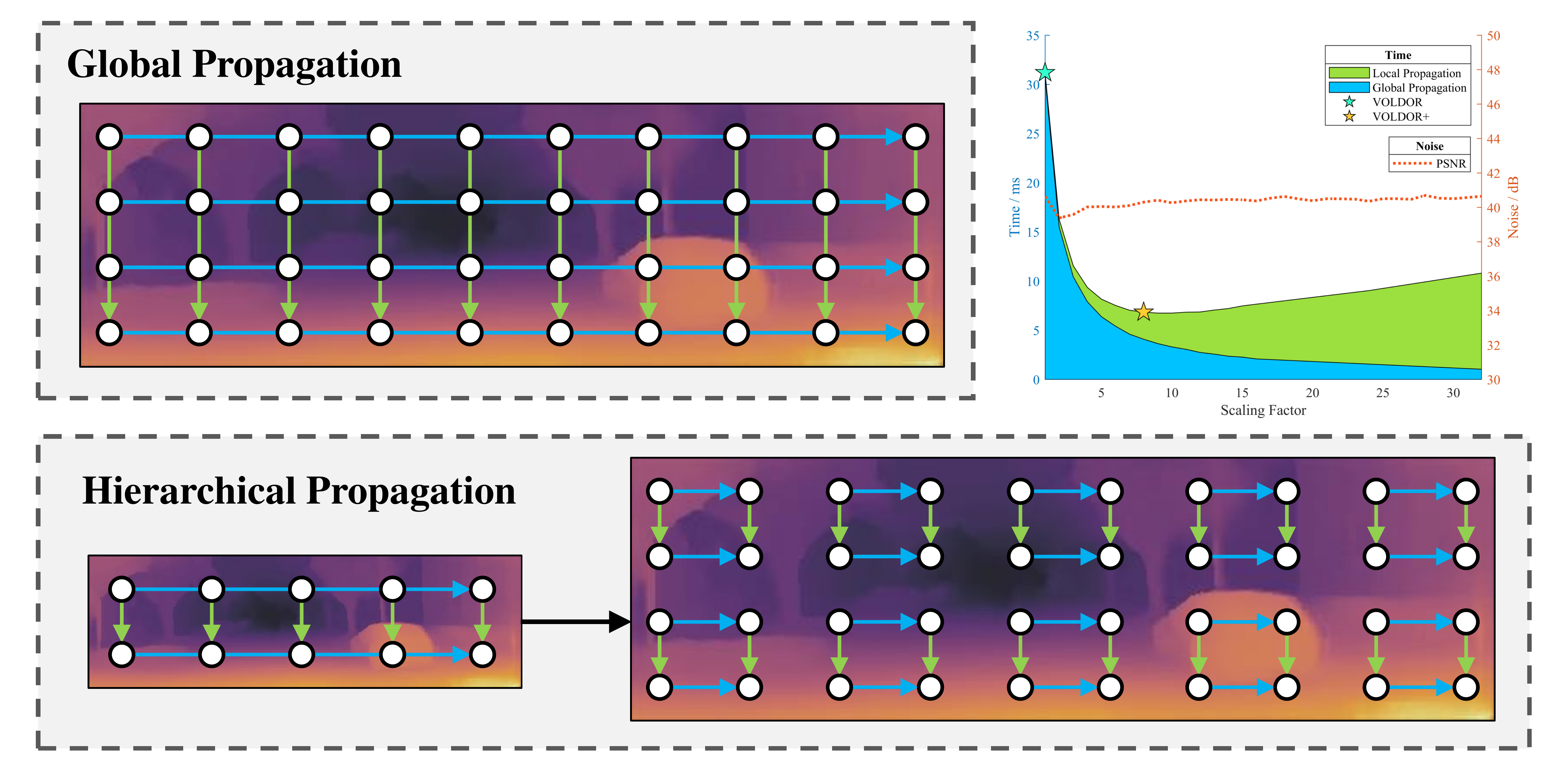}
\caption{Depth propagation workflow. 
Top right shows runtime (ms) vs depth map  quality (PSNR) trade-offs of our hierarchical  propagation scheme across varying scaling factors.}
\label{fig:depth_propagation}
\end{figure}

 \noindent \textbf{Hierarchical Propagation Scheme}.
For estimating  depth  and rigidness maps, \cite{min2020voldor} adopts a sampling and propagation scheme \cite{zheng2014patchmatch} based on a generalized-EM framework. The image 2D field is broken into alternatively directed 1D chains. In the M-step, depth values are randomly sampled and propagated through each chain. In the E-step, hidden Markov chain smoothing is imposed on the chains of rigidness maps using the forward-backward algorithm.  
Such sequential global depth propagation scheme imposes a  performance bottleneck for GPU implementations, as it is not massively parallelizable. Hence, VOLDOR$^+$ proposes a hierarchical propagation scheme, where a global propagation is done on a depth map of reduced scale to ensure convergence rate, and  local propagations  done in local windows at full resolution to retain fine details. Per Fig.\ref{fig:depth_propagation}, VOLDOR$^+$ achieves 3x speed-up  w/o loss in depth map quality w.r.t. \cite{min2020voldor}.

\noindent \textbf{VO Stride and Keyframe Selection.} 
Once VOLDOR$^+$ runs on an input batch and estimates local 3D geometries,
we determine the stride to the next batch and select keyframes using visibility-coverage (VC) metrics. The VC metrics are defined over a pair of frames $I_1, I_2$ with known poses, where only $I_1$ may have a depth map.
 We define  visibility score as the proportion of depth pixels in $I_1$ whose projection in $I_2$ is inside $I_2$'s image frame. Conversely, we define the coverage score as the proportion of the image area in $I_2$ being covered by the depth pixels' projection from $I_1$. 
We define the VC score as the harmonic mean of visibility and coverage
\begin{equation}
    VC = \frac{2}{{Visibility}^{-1}+{Coverage}^{-1}}
\end{equation}
The VC score is estimated between the reference frame and  all other frames within the batch. The first frame with VC score below $\tau_{stride}$ is selected as the next batch's reference.
Finally, if the VC score between the current batch's reference frame and the latest keyframe is below $\tau_{keyframe}$, we register the current batch's reference frame as a keyframe.



\subsection{Back-End}
The front-end VO only establishes local pair-wise constraints between successive frames $I_t$ and $I_{t+1}$ within an input batch. Conversely, the back-end manages spatial relationships of larger scope by establishing spatial constraints between keyframes and detecting loop closure instances.

\noindent \textbf{Keyframe Alignment.}
We establish keyframe alignment (KA) constraints by 3D registration of their depthmaps. 
That is,  for two depth maps $\boldsymbol{\theta_1}$ and $\boldsymbol{\theta_2}$, we estimate both their relative motion $\boldsymbol{T}$ and scale factor $s$.
Similar to \cite{park2017colored}, we formulate the depth image alignment objective in terms of the difference of the inverse depth values
\begin{equation} \label{Eq_geo_slow}
    E_{geo} = \sum_{j} C^j \, \psi \! 
    \left(
    \frac{1}{s\theta_1^{\ell}}
    - 
    \frac{1}{\phi_z(\theta_2^j, \boldsymbol{T})}
    \right)
\end{equation}
where $\psi(\cdot)$ is the Cauchy kernel function,  $C^j$ is the confidence associated with pixel $j$ defined in Eq.\eqref{eq:depth_conficence}.
While effective, the convergence rate of Eq.\eqref{Eq_geo_slow} impeded online operation. Towards this end, we generalize the point-to-plane error to account for an inverse depth parameterization:
\begin{equation}
        E_{geo} = \sum_{j} C^j \, \psi \!  \left(  
        \frac{ \langle 
        \mathbf{n}\left({\theta}_1^{\ell}\right) 
        , \;
        \phi ( \theta_1^{\ell}, \boldsymbol{I}) \!-\! \phi(\theta_2^j, \boldsymbol{T})
        \rangle }
        {s  \theta_1^{\ell}  \phi_z(\theta_2^j,\boldsymbol{T})}
        \right)
\end{equation}
where $\langle \cdot, \cdot \rangle$ denotes the vector inner product, while $\mathbf{n}({\theta_i^j})$ denotes the normal vector at pixel $j$ of depth map $\boldsymbol{\theta}_i$. Such objective
implicitly enforces spatial regularization given that $\mathbf{n}({\theta_i^j})$ is computed from the local
 depth estimates and significantly increases convergence speed and quality.
Geometrically,  scaling pair-wise point-to-plane distances among depth estimates $\theta^\ell_1$ and  $\theta^j_2$ inversely proportional to their depths,  prioritizes geometry nearby to the reference frame. Conversely, weighting each kernel output by our confidence measure, mitigates  depth outliers.
  Optionally, our keyframe alignment process can benefit from the input intensity images by adding the energy function for photometric consistency:
\begin{equation} \label{eq:frame_alignment_photo_consistency}
        E_{photo} = \sum_{j} C^j \, \psi \! \left(  (I_1^\ell - b_1) - \frac{e^{a_1}}{e^{a_2}}(I_2^j-b_2) \right)
\end{equation}
where $a_1,a_2,b_1,b_2$ are the pixel brightness affine transformation parameters to be estimated, while $I_i^j$ yields the intensity value of pixel $j$ in image $I_i$. The photometric term is optional since we favor our  system to be an indirect method, and rely on external \OF{} modules to handle common intensity-based imaging aberrations such as exposure and/or white-balance variations, non-Lambertian reflections, etc.
Also,  VOLDOR$^+$ estimates depth maps from small batches of successive imagery, while temporally distant observations are subject to arbitrary changes in appearance (i.e. global illumination, shadows, specularities) even if their underlying geometry is consistent.
Moreover,  our depth inference framework targets the estimation of static geometry observed throughout a multi-frame input batch. Empirically, we observe it consistently "looks through" dynamic or small foreground structures observed in the batch (first) reference frame and estimates  the background's depth.
For such cases, the photometric term may reduce the overall accuracy. 
Finally, the criteria for keyframe alignment is
\begin{equation}
\label{Eq_KA}
    \{{\boldsymbol{T},s}\} = \underset{\boldsymbol{T}^*,s^*}{argmin} (E_{geo} + \lambda \cdot E_{photo})
\end{equation}
where $\lambda$  balances  geometry and photometric errors, which are omitted when intensity images are absent. The relative depth scaling factor $s$ is set to $1$ for depth input with absolute scale (e.g. stereo or RGB-D). Once Eq.\eqref{Eq_KA} is optimized using Levenberg-Marquardt, the covariance of  $\boldsymbol{T}$ is linearly approximated through its Jacobian matrix as $\boldsymbol{\Sigma}=(\boldsymbol{J}^\top \boldsymbol{J})^{-1}$.\\
\noindent \textbf{Priority Linking.} The aggregation of all pairwise geometric constraints among keyframes yields a fully connected graph (i.e. quadratic  complexity). To enable online operation, we avoid exhaustive exploration of these relationships by devising an adaptive prioritization mechanism aimed at: 
1) Linking keyframes with sufficient overlap for robust and accurate depth map 3D alignment
2) Balancing exploration for potential loop closures vs. exploiting local connectivity.
While loop closure benefits are self-evident, strengthening local connectivity fosters pose corrections to the current keyframe, which may be used as geometric priors in the VOLDOR$^+$ front-end, contributing to more accurate VO estimates.
 We construct a priority matrix $\mathbf{Q}$ for pairwise keyframe linkage, based on the observation that temporal proximity serves as a low-cost proxy for VC scores.
  That is, the distance between indexes roughly encodes co-visibility under the assumption of smooth local  motion.
Accordingly, we systematically update the entries in $\mathbf{Q}$ 
based on two linkage types (i.e. realtime linkage and loop closure linkage).
First, for realtime linkages, after the current keyframe with index $\kappa$ is created, we update the priority of each arbitrary keyframe pair with indices $i,j$, as 
\begin{equation} \label{eq:priority_local}
    \dot{Q}_{i,j} = exp \left( -\frac{(i-j)^2}{\sigma_{spatial}^2} - \frac{(\kappa-i)(\kappa-j)}{\sigma_{temporal}^2} \right),
\end{equation}
 where the parameter  $\sigma_{spatial}$  controls the priority w.r.t. co-visibility (proximal keyframes correlate to larger shared overlaps), while $\sigma_{temporal}$  controls the priority w.r.t. the timeliness of considered keyframe pair (we prioritize recent keyframes over older ones).
Second, to foster the inclusion of loop closure linkages, we run an image retrieval engine based on the 
DBoW3 \cite{GalvezTRO12} library, using the current keyframe $\kappa$ as query, to obtain a candidate keyframe $\kappa'$. 
Assume $\mathcal{L}^{th}$ pair $(\kappa,\kappa')$ passes a geometric verification test, 
whenever $| i - \kappa | \leq | j- \kappa | $,
we compute the linking priorities in the proximity of the detected overlap as:
\begin{equation} \label{eq:priority_lc}
    Q_{i,j}^{\mathcal{L}} = Q_{j,i}^{\mathcal{L}} = exp \left( -\frac{\left( | i - \kappa | + | j- \kappa' | \right)^2}{\sigma_{lc}^2} \right)
\end{equation}
where $\sigma_{lc}$ controls the scope of priority propagation surrounding the loop closure pair. Given $N_{lc}$ total detected loop closure pairs, the final linking priority matrix $\boldsymbol{Q}$ is
\begin{equation}
    Q_{i,j}=max \left( \dot{Q}_{i,j}, \; max(Q_{i,j}^\mathcal{L} \mid \mathcal{L}=1 \dots N_{lc}) \right)
\end{equation}
The largest unlinked element in $\mathbf Q$ larger than $\tau_{link}$ will be processed by the keyframe alignment module to determine an accurate relative motion among their associated depth maps. Finally, the selected pairwise motion constraint is incorporated into a  $sim(3)$ pose graph framework \cite{strasdat2010scale}, which enforces both globally consistent correction propagation (i.e. from loop closure linkages) as well as local refinements (i.e. from realtime linkages).

\begin{table*}[t!] 
\centering
\resizebox{2\columnwidth}{!}{
\begin{tabular}{|c|cc|cc|cc|cc|cc|cc|}
\hline
 & \multicolumn{6}{c|}{Stereo} & \multicolumn{6}{c|}{Monocular} \\ \cline{2-13} 
\begin{tabular}[c]{@{}c@{}}Sequence\\ (Hard Only)\end{tabular} & \multicolumn{2}{c|}{ORB-SLAM3} & \multicolumn{2}{c|}{\begin{tabular}[c]{@{}c@{}}VOLDOR$^+$\end{tabular}} & \multicolumn{2}{c|}{VOLDOR$^+$SLAM} & \multicolumn{2}{c|}{ORB-SLAM3} & \multicolumn{2}{c|}{DSO} & \multicolumn{2}{c|}{VOLDOR$^+$SLAM} \\ \cline{2-13} 
 & \begin{tabular}[c]{@{}c@{}}Trans.\\ (\%)\end{tabular} & \begin{tabular}[c]{@{}c@{}}Rot.\\ (deg/m)\end{tabular} & \begin{tabular}[c]{@{}c@{}}Trans.\\ (\%)\end{tabular} & \begin{tabular}[c]{@{}c@{}}Rot.\\ (deg/m)\end{tabular} & \begin{tabular}[c]{@{}c@{}}Trans.\\ (\%)\end{tabular} & \begin{tabular}[c]{@{}c@{}}Rot.\\ (deg/m)\end{tabular} & \begin{tabular}[c]{@{}c@{}}Comp.\\ (\%)\end{tabular} & \begin{tabular}[c]{@{}c@{}}Rot.\\ (deg/m)\end{tabular} & \begin{tabular}[c]{@{}c@{}}Comp.\\ (\%)\end{tabular} & \begin{tabular}[c]{@{}c@{}}Rot.\\ (deg/m)\end{tabular} & \begin{tabular}[c]{@{}c@{}}Comp.\\ (\%)\end{tabular} & \begin{tabular}[c]{@{}c@{}}Rot.\\ (deg/m)\end{tabular} \\ \hline
abandonedfactory & 0.0404 & \textbf{0.0914} & 0.0345 & 0.1095 & \textbf{0.0269} & 0.0924 & 0.969 & \textbf{0.2204} & 0.970 & 0.2624 & \textbf{0.983} & 0.3210 \\
abandonedfactory\_night & 0.1633 & 0.0765 & 0.0227 & 0.0800 & \textbf{0.0196} & \textbf{0.0702} & 0.932 & 0.4907 & 0.916 & 1.4233 & \textbf{0.964} & \textbf{0.2229} \\
amusement & 0.0129 & 0.0654 & 0.0125 & 0.0315 & \textbf{0.0118} & \textbf{0.0265} & 0.640 & - & 0.505 & - & \textbf{1.000} & \textbf{0.1188} \\
carwelding & \textbf{0.0060} & \textbf{0.0269} & 0.0124 & 0.0518 & 0.0131 & 0.0535 & 0.995 & \textbf{0.0850} & 0.435 & - & \textbf{0.997} & 0.3203 \\
ocean & 0.0664 & 0.4379 & 0.0376 & 0.1340 & \textbf{0.0335} & \textbf{0.1311} & 0.908 & 0.7564 & 0.898 & 1.7069 & \textbf{0.958} & \textbf{0.4775} \\
office & \textbf{0.0035} & \textbf{0.0323} & 0.0088 & 0.0740 & 0.0071 & 0.0624 & 0.907 & 0.2872 & \textbf{0.950} & 4.4242 & 0.853 & \textbf{0.1715} \\
japanesealley & 0.0193 & 0.0773 & 0.0105 & \textbf{0.0369} & \textbf{0.0102} & 0.0413 & 0.964 & 0.1819 & 0.959 & 0.1291 & \textbf{1.000} & \textbf{0.1150} \\
seasonsforest & 0.1160 & 0.3102 & 0.0293 & 0.1094 & \textbf{0.0196} & \textbf{0.0733} & 0.330 & - & 0.534 & - & \textbf{1.000} & \textbf{0.1525} \\
westerndesert & 0.0906 & 0.3311 & 0.0177 & 0.0898 & \textbf{0.0149} & \textbf{0.0786} & 0.918 & 0.3933 & 0.855 & 1.4918 & \textbf{0.957} & \textbf{0.2145} \\ \hline
Avg. & 0.0576 & 0.1610 & 0.0207 & 0.0797 & \textbf{0.0174} & \textbf{0.0699} & 0.840 & 0.3449 & 0.780 & 1.5729 & \textbf{0.968} & \textbf{0.2348} \\ \hline
\end{tabular}
}
\caption{Camera pose accuracy evaluated on TartanAir dataset. Translation and rotation error metrics are based on normalized relative pose error as in KITTI \cite{kitt2010visual}, but averaged over all possible sub-sequences of length $(5, 10, 15, \dots ,40)$ meters. The completeness metric measures the number of successfully registered frames, which  reveals a method's robustness under each environment. If the completeness of a certain sequence falls bellow 0.8, we do not estimate  rotation error.}\label{table:pose_accuracy}
\end{table*}

\begin{figure}[]
\centering
\includegraphics[width=0.492\columnwidth]{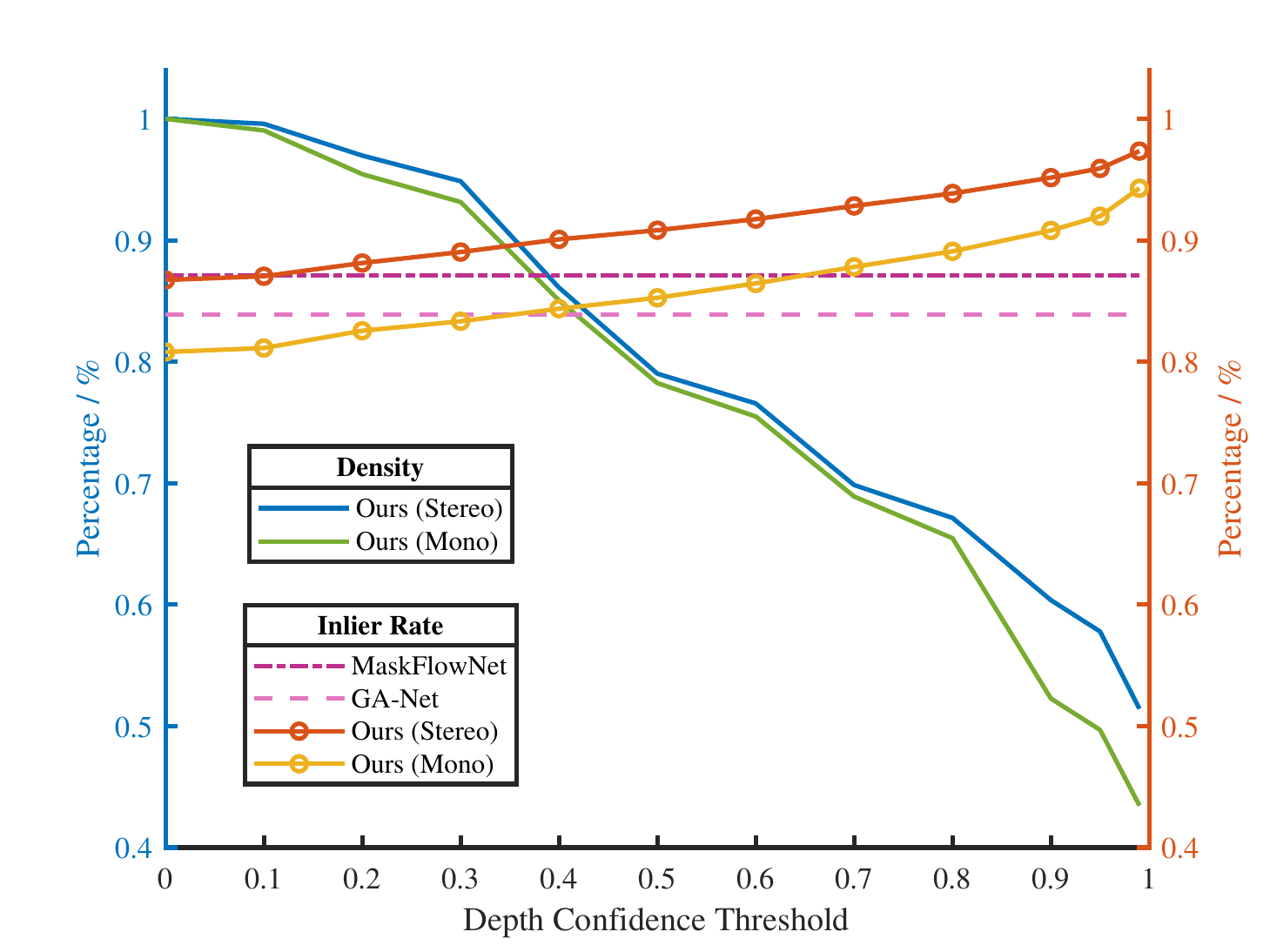}
\includegraphics[width=0.492\columnwidth]{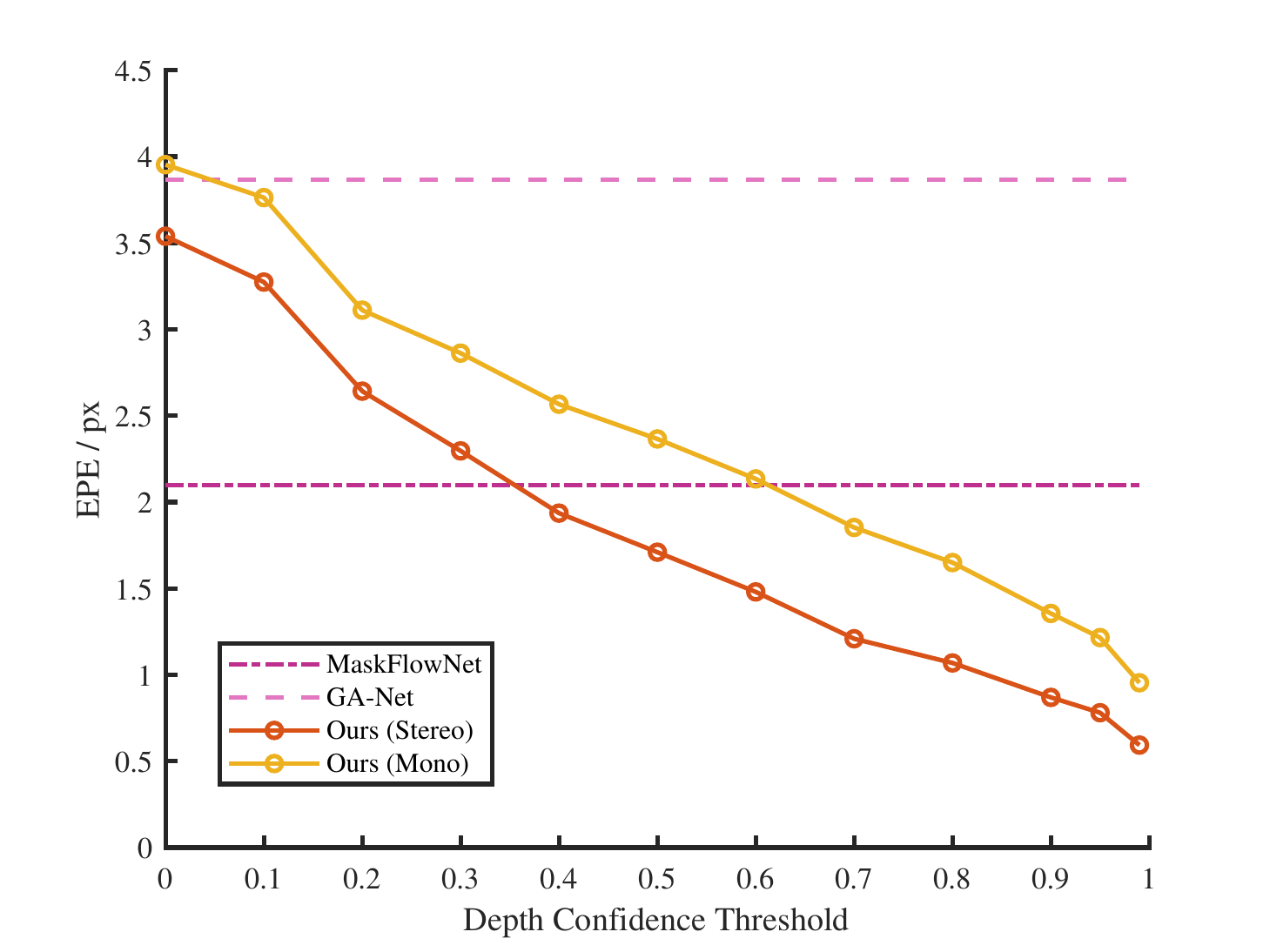}
\caption{Depth map quality evaluated on TartanAir dataset. We threshold over depth confidence value to evaluate pixel density, inlier rate and EPE across different confidence levels.}
\label{fig:depth_accuracy}
\end{figure}

\begin{figure*}[b!]
\centering
\includegraphics[width=1.92\columnwidth]{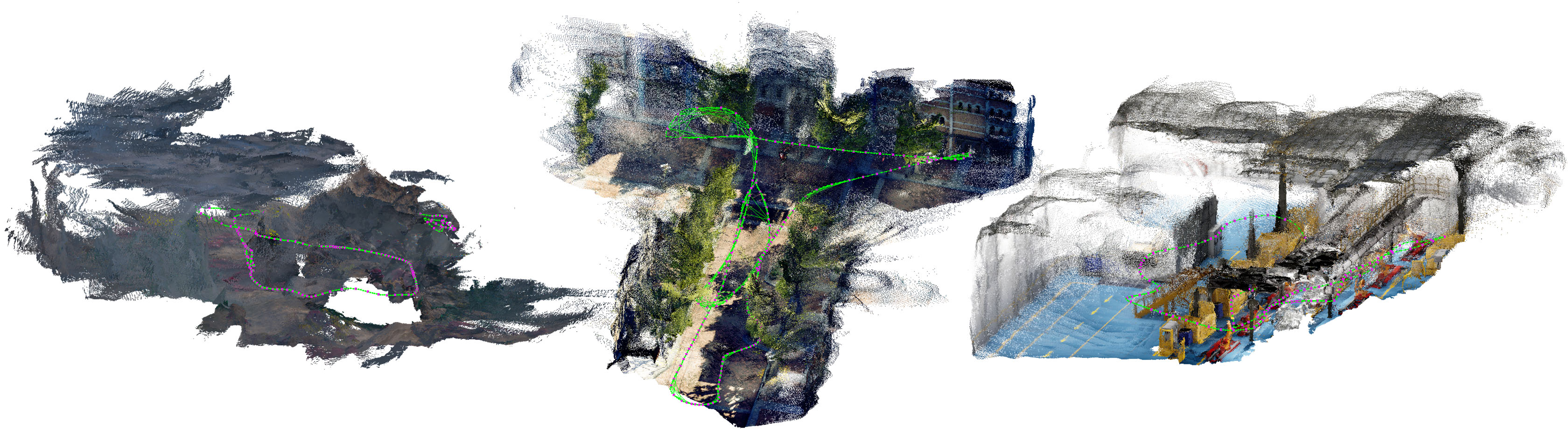}
\text{(a) Results on TartanAir 'ocean', 'oldtown' and 'carwelding' sequence with stereo input.}
\\
\includegraphics[width=0.58\columnwidth]{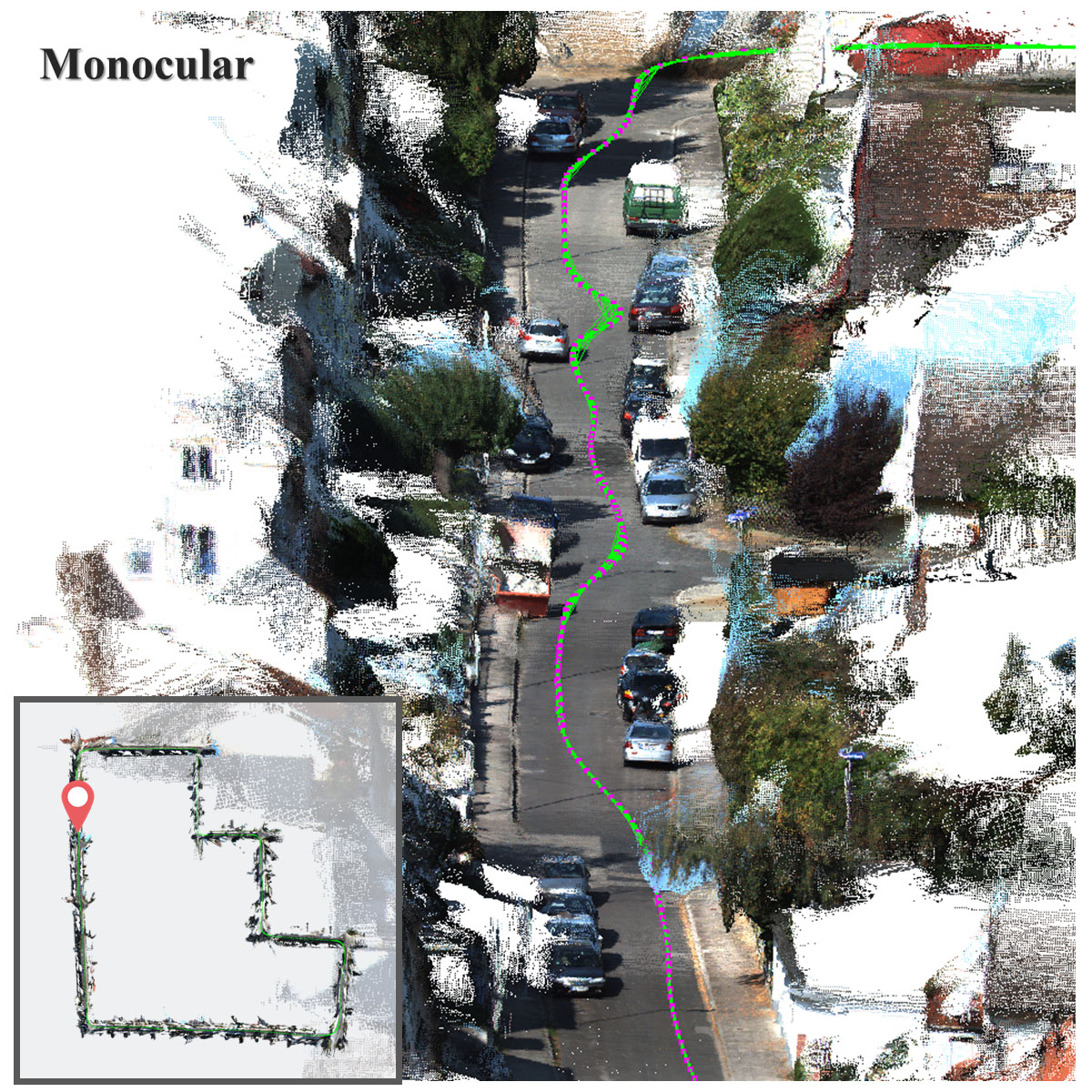}
\;\;\;
\includegraphics[width=0.58\columnwidth]{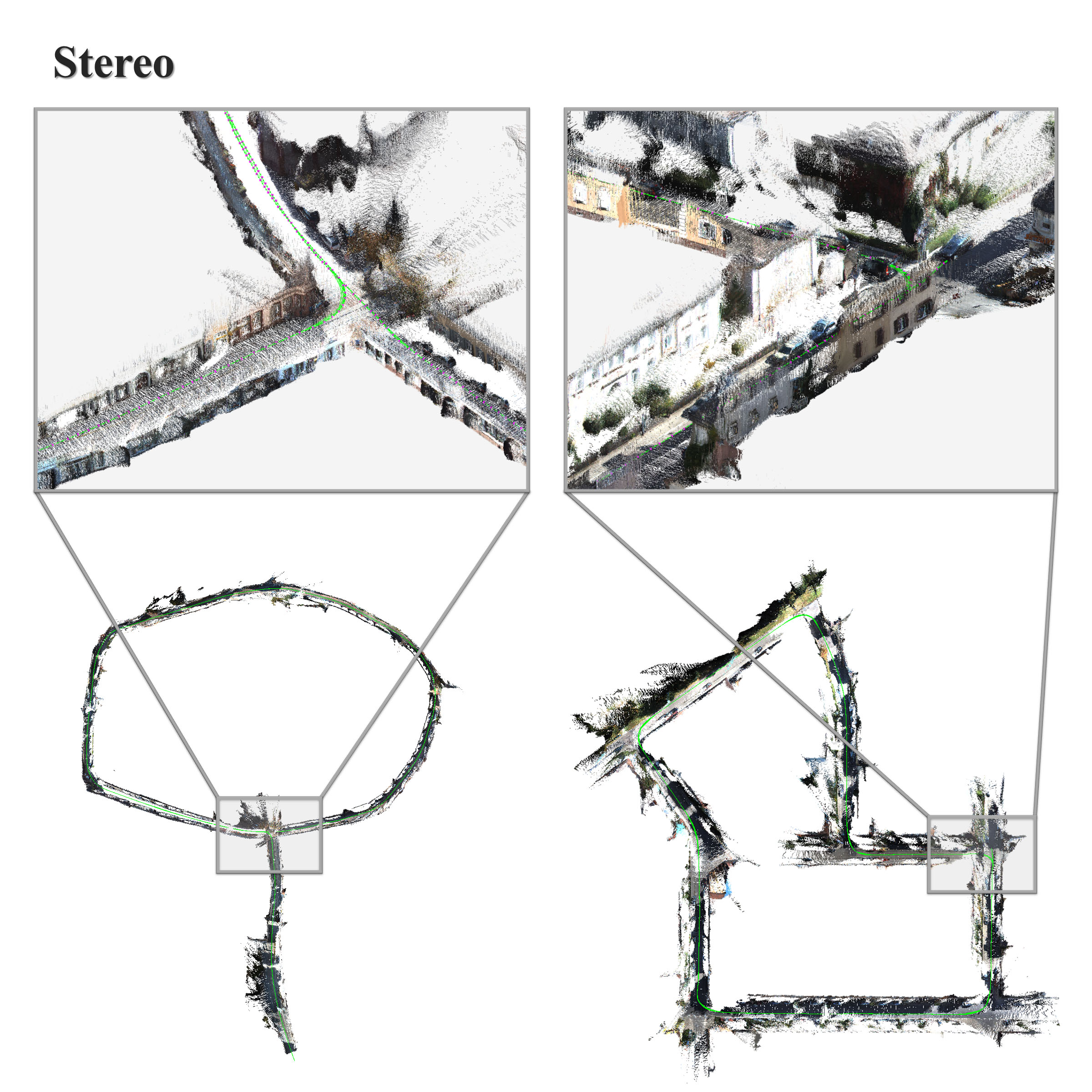}
\;\;\;
\includegraphics[width=0.58\columnwidth]{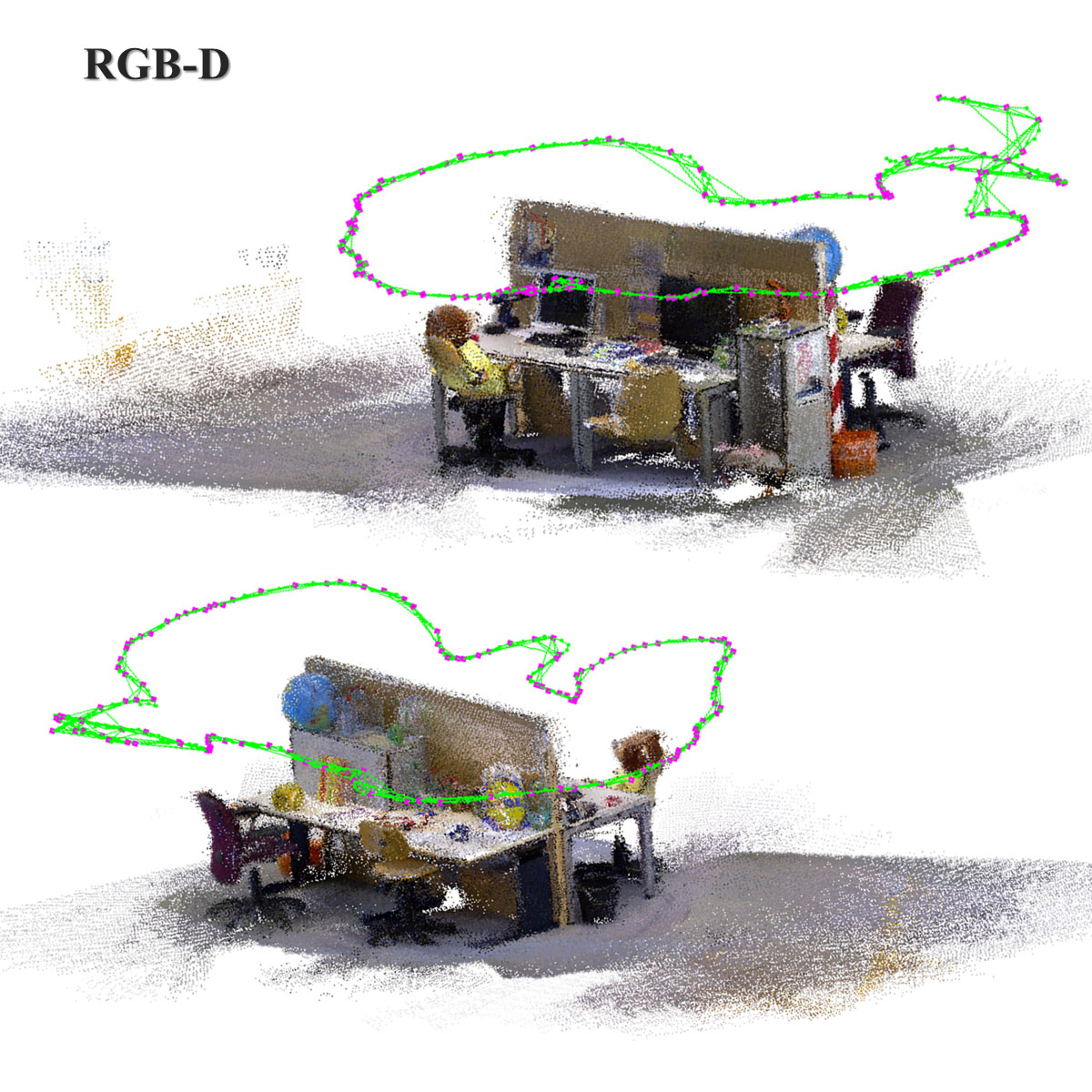}
\text{(b) Results  on KITTI and TUM RGB-D datasets with different types of input. }
\caption{Qualitative Results. Top: Results across diverse environments show VOLDOR$^+$SLAM's  robustness to capture variability. Bottom: Zoomed-in  regions and global views illustrate our 3D mapping density, consistency and level of detail.}
\label{fig:qualitative_results}
\end{figure*}

\section{Experiments}
\noindent{\bf Experimental Setup}. 
We benchmarked on the photo-realistic synthetic dataset TartanAir\cite{wang2020tartanair}, which provides ground-truth camera poses, depth maps and optical flows. The dataset features challenging environments in the presence of moving objects, changing light, various weather conditions,  with diverse viewpoints and motion patterns that are difficult to obtain in real world. We tested on the "hard" track of 9 sequences covering diverse indoor and outdoor environments. 
Our {\em off-the-shelf} \OF{} estimator MaskFlowNet \cite{zhao2020maskflownet} was not trained or fine-tuned on the TartanAir dataset.
For stereo input, we use the X-component of the \OF{} estimated from the left to the right camera as the disparity map, enabling us to use the same empirical residual model as described in \cite{min2020voldor} both for stereo and optical flow.
 For all sequences, we use the photometric consistency term of Eq.\eqref{eq:frame_alignment_photo_consistency}. We compared VOLDOR$^+$SLAM (full pipeline) and VOLDOR$^+$ (VO-only) vs.  ORB-SLAM3 \cite{campos2020orb} (stereo and monocular) and DSO \cite{engel2017direct} (monocular). 
 \\
\noindent{\bf Stereo Results}. 
 Per Table.\ref{table:pose_accuracy} (left)  ORB-SLAM3 excels on 
 stable environments with enough textures such as 'office' and 'carwelding', but accuracy drops dramatically for sequences exhibiting poorly-textured regions and rapidly changing viewpoints (i.e. trees that rapidly move closer) such as 'ocean' and 'seasonsforest'. Conversely, both our variants perform stably across various environments with considerable improvement from our full  pipeline, which has the best overall pose accuracy scores. \\
\noindent{\bf Monocular Results}.
For monocular input, the challenging dataset caused our  baselines to frequently lose tracking, obfuscating a fair comparison over translation error, which requires scale correction due to scale ambiguity. That is, whenever a system loses tracking, a new map with arbitrary scale is generated. If we scale each sub-sequence respectively, the system losing tracking more often will benefit from getting more accurate scale correction. 
Thus, we replace translation error with the completeness metric, similar to the success rate proposed in \cite{wang2020tartanair}.
  The results in  Table.\ref{table:pose_accuracy} (right) show VOLDOR$^+$SLAM can robustly handle different environments with high completeness compared to our baselines, while achieving best overall rotation accuracy. \\
\noindent{\bf Depth Evaluation}.
Fig.\ref{fig:depth_accuracy} conveys our depth map quality results.
As baselines, we use  GA-Net \cite{Zhang_2019_CVPR} and MaskFlowNet \cite{zhao2020maskflownet} (the stereo input of our stereo-based VO). The accuracy metrics used are inlier-rate and EPE. A depth value is considered inlier when its corresponding disparity EPE is less than $3$ pixels or $5\%$ ground-truth disparity.
Results show  our subset of pixels with high confidence greatly out-performs existing baselines, where around $50\%$ pixels have a confidence over $0.99$. Our framework provides only slightly more accurate depth maps using stereo instead of monocular input.
Since VOLDOR$^+$ 
disadvantages the estimation of observed dynamic content,
 we deem  reported accuracy values an under-estimate for static environments.\\
 \noindent{\bf Qualitative Results.} Representative reconstructions from TartanAir, KITTI and TUM datasets can be found in Fig.\ref{fig:qualitative_results}.

\section{Conclusions}
VOLDOR$^+$SLAM demonstrates the potential of dense-indirect (DI) estimation frameworks for the geometric analysis of large-scale and unstructured environments.
The modular nature of the DI approach allows for the exploration of novel formulations and ancillary tools. In particular, recently proposed learning approaches jointly estimating depth and poses \cite{zhou2017unsupervised,yin2018geonet,wang2019recurrent,yang2020d3vo}, and multi-way 3D registrations \cite{lv2019taking,li2018deepim,zhou2018deeptam,tang2018ba} are highly related to our work and offer promising research paths toward tighter couplings between DI estimation and global map management modules.




\bibliographystyle{IEEEtran}
\bibliography{mybib}

\end{document}